\documentclass[letterpaper, 10 pt, conference]{ieeeconf}  %

\IEEEoverridecommandlockouts                              %

\usepackage{graphics} %
\usepackage{epsfig} %
\usepackage{times} %
\usepackage{todonotes}
\usepackage{amsmath, amssymb, amsfonts}
\usepackage{subcaption}
\usepackage{float}
\usepackage{multirow}
\usepackage{wrapfig}

\usepackage[backend=bibtex,style=ieee,natbib=true,maxcitenames=1,mincitenames=1]{biblatex} %
\usepackage[nolist]{acronym}
\usepackage{hyperref}

\DeclareMathOperator*{\E}{\mathbb{E}}
\newcommand{\e}[2]{\E_{#1} \left[ #2 \right]}
\newcommand{\longe}[2]{\hspace{-75pt}\E_{\hspace{80pt} #1}\hspace{-75pt} \left[ #2 \right]}

\title{\LARGE \bf
Multimodal dynamics modeling for off-road autonomous vehicles
}

\author{Jean-Fran\c{c}ois Tremblay$^{1}$, Travis Manderson$^{1}$, Aur\'elio Noca$^{1}$, Gregory Dudek$^{1}$ and David Meger$^{1}$%
\thanks{$^1$All authors are with the Mobile Robotics Laboratory (MRL), McGill University, Montr\'eal, Canada, H3A 2A7. Corresponding e-mail: \texttt{jft@cim.mcgill.ca}
This work was supported by the Fonds de Recherche du Qu\'ebec, Nature et Technologie (FRQNT), Hydro-Qu\'ebec and the NSERC Canadian Robotics Network (NCRN).}
}

\begin{document}

\maketitle
\thispagestyle{empty}
\pagestyle{empty}

\begin{abstract}
Dynamics modeling in outdoor and unstructured environments is difficult because different elements in the environment interact with the robot in ways that can be hard to predict.
Leveraging multiple sensors to perceive maximal information about the robot's environment is thus crucial when building a model to perform predictions about the robot's dynamics with the goal of doing motion planning.
We design a model capable of long-horizon motion predictions, leveraging vision, lidar and proprioception, which is robust to arbitrarily missing modalities at test time.
We demonstrate in simulation that our model is able to leverage vision to predict traction changes.
We then test our model using a real-world challenging dataset of a robot navigating through a forest, performing predictions in trajectories unseen during training.
We try different modality combinations at test time and show that, while our model performs best when all modalities are present, it is still able to perform better than the baseline even when receiving only raw vision input and no proprioception, as well as when only receiving proprioception.
Overall, our study demonstrates the importance of leveraging multiple sensors when doing dynamics modeling in outdoor conditions. 

\end{abstract}

\begin{acronym}
\acro{EKF}{extended Kalman filter}
\acro{IMU}{inertial measurement unit}
\acro{ANN}{artificial neural network}
\acro{CNN}{convolutional neural network}
\acro{RNN}{recurrent neural network}
\acro{GRU}{gated recurrent unit}
\acro{SLAM}{simultaneous localization and mapping}
\acro{MBRL}{model-based reinforcement learning}
\acro{RL}{reinforcement learning}
\acro{RSSM}{recurrent state-space model}
\acro{MVAE}{multimodal variational autoencoder}
\acro{VAE}{variational autoencoder}
\acro{RSSM}{recurrent state-space model}
\acro{SLAC}{stochastic latent actor-critic}
\acro{ICP}{iterative closest point}
\acro{ELBO}{evidence lower bound}
\acro{SAC}{soft actor-critic}
\acro{RMSE}{root mean square error}
\acro{MRSSM}{multimodal recurrent state-space model}
\end{acronym}

\section{Introduction}

A wheeled robot moving in the forest interacts with rocks, trees, ice, mud, and a multitude of other elements, making its motions complex and difficult to predict -- a fact known to generations of humans who have previously been the pilots of off-road vehicles.
The prediction of such future motion is key to control autonomous off-road vehicles. %
Understanding likely physical outcomes can begin with knowing the proprioceptive and command information, which are highly predictive on human-made roads, but exteroception is critical to predict motions and to reason about roughness, ground traction or slipping, and contact with foliage. Furthermore, each sensor in this environment should be treated as unreliable, as they can fail from a variety of causes, from low light to being covered with mud and soil. 

This paper proposes a novel dynamics modeling approach suited to motion prediction in the forest.
We extend a recent multimodal, latent space approach for dynamics modeling that is robust to missing modalities. Our method scales well with the number of sensors present and is able to model the vehicle's dynamics from different sensor combinations, taking advantage of multiple exteroceptive sensors when available.
These predictions are propagated through time building upon a \ac{RNN} based model \citep{planet} enabling prediction over long horizons.

Our latent embedding allows us to build an abstract representation of the state of the robot in a self-supervised manner, which automatically incorporates information such as terrain classes and obstacles such as tree branches. We learn without any manual terrain labeling, which would be challenging in our scenario: while it is possible to label the few classes present in urban driving (road, sidewalk, grass, etc) there are much wider range of less clearly demarcated terrain elements present in the forest (Is the robot on a rock? Which wheel is on the rock? Is the rock wet?) Rather, our method uses \emph{post hoc} state-estimation from on-board vehicle lidar \citep{francoisicpmapper} as a direct training signal, enabling future prediction for several seconds forward in time, in new trajectories in the same forest region. 

Multimodality enables our model to leverage all sensors present on the robot, giving it access to maximal information about the state of the robot and its surrounding environment.
However, sensors can be prone to failure, especially in natural and remote environments where replacements or repair are not immediate
Thus, we design our dynamics model to be able to perform predictions without retraining and with minimal loss of accuracy in the case of missing modalities.
For example, a forest robot can benefit from the combination of vision, lidar and \acp{IMU} during the daytime and seamlessly fall back to lidar and \acp{IMU} only during the nighttime with limited impact on the performance.
Additionally, our approach can accommodate sensors with different rates where one or more sensors are effectively unavailable at a given timestep.

We demonstrate the accuracy of our method, in terms of the final pose predictions, on our real-world forest dataset, where we have 17 trajectories of a robot traversing forest terrain. Our results show that our model is able to benefit from all modalities when they are present, performing better than our baseline, while still handling missing data when only receiving raw vision or proprioception.

Overall, our contributions include:
\begin{itemize}
    \item The derivation of our \ac{MRSSM}, an approach that combines multimodal learning with latent time-series prediction.  
    \item The derivation of a new \ac{MVAE} \ac{ELBO} that maximizes the likelihood of held-out modalities and takes into account all available training data. 
    \item The first use, to our knowledge, of \acp{MVAE} for challenging outdoor robotics data.
    \item Demonstration of missing-modality prediction robustness, again within this challenging domain.
\end{itemize}

\section{Related work}
\label{sec:relworks}
We first review our application domain, autonomous navigation in outdoor environments, and then move to the related algorithms for model-based reinforcement learning and multimodal learning.

\paragraph{Navigation in difficult outdoor environments}
Autonomous robots navigating in unstructured, outdoor environments face adverse conditions in terms of complex, uneven terrain that is often slippery and contains rubble that causes slippage. Additionally, vegetation often intermittently occludes sensors such as lidar and cameras. A common strategy for learning where to navigate in these environments is to use behavior cloning~\cite{pomerleau1989alvinn} where a user provides demonstrations of traversing natural paths~\cite{Giusti2016, Smolyanskiy2017}.
Similarly, end-to-end techniques have been used to effectively learn a navigation policy by providing expert steering demonstrations for autonomous driving~\cite{bojarski_nvidia_driving} as well as difficult underwater environments~\cite{Manderson2018, Manderson-RSS-20}.  
Another common approach is to perform appearance-based segmentation, and train a classifier to label the segmentation classes with a score corresponding to the navigability difficulty~\cite{Sofman2006, Rothrock2016, Wang2018, Delmerico2017}.
Often, learning the navigability score is difficult, so some authors~\cite{Barnes2017, Tang2017} have used expert demonstrations to learn the traversable classes.

\paragraph{\Ac{MBRL}} is an approach to \ac{RL} where we learn to predict the results of our action on the state of the robot and the reward through \mbox{(self-)supervised} learning.
This is usually viewed as a regression problem, which can be solved though Gaussian processes, for example \citep{ko2009gp}, \citep{pilco}.
Neural networks can also be used \citep{juan}.
Until recently, these models did not work on high-dimensional observations such as vision.
Visual foresight \citep{visualforesightjournal, visualforesighticra, hierarchicalforesight, videoprediction} was developed in a series of papers and used a video prediction model to learn visual dynamics.

\paragraph{Latent space models} are one category of approaches for \ac{MBRL}.
They rely on mapping observations to a latent space, usually of lower dimensionality than the observation space, and perform prediction in that space.
This dimensionality reduction helps sample efficiency and is learned, usually with reconstruction losses using some encoder-decoder combination
\citep{planet,deepmdp,worldmodels}.
This abstract latent space, in our case, is useful for fusing different observation modalities.

\paragraph{Multimodal learning}
There are multiple facets contributing to the utility of multimodal learning algorithms, the most important being: their scalability to a high number of modalities, their robustness to missing modalities as well as their performance on the task at hand achieved by taking advantage of all of the modalities present.
\citet{travis} performed the fusion of drone and ground-robot imagery, in the context of robot dynamics modeling, by concatenating the output of two convolutional encoders and feeding the result into a \ac{RNN}.
\citet{grfu} proposed Gated Recurrent Fusion Units to perform weighted fusion of multimodal data in the context of self-driving vehicles.
Their approach allows the model to give more importance to different modalities through weighting, while our approach does the same through uncertainty-aware fusion using a product of experts.
However, their approach does not allow for missing modalities.
Using a different philosophy altogether, \citet{nyumultimodal} described an approach to perform sensor selection for a reactive steering policy.
Their approach did not perform fusion but rather dynamically selected one sensor from four cameras and one lidar to perform the prediction.
However, we argue that leveraging all available information rather than choosing one particular sensor is a better choice in most cases.
Closer to \citep{mvae}, detailed in \autoref{sec:mvae}, \citet{wrongmvae} develop another approach to multimodal variational autoencoder, effectively setting all missing modalities values to -2 randomly at training time. While this approach is capable of representing data with many modalities, the approach of \citet{mvae} has shown the ideal combination of flexibility to missing modalities and prediction accuracy, which we argue is crucial to deploy multimodal models on field robots.

\section{Methods}
\label{sec:methods}

We seek a predictive dynamics model capable of capturing the motions of a robot over several seconds through environments with complex wheel-terrain-obstacle interactions.
High-dimensional sensing with images and/or lidar allows interpretation of the environment, and we therefore select generative time-series models that can interpret and predict physical robot states as well as images.
In particular, we turn to the so-called \ac{RSSM} introduced in \citep{planet}, for its ability to accurately model transitions in high-dimensional sensory spaces. A \ac{RSSM} is a time-series latent variable model, where observations $o_t$ at time $t$, do not directly contain all relevant information, such as wheel contact and environment structure. Therefore, our method will attempt to recover a latent state representation $s_t$ sufficient to explain observations and, along with actions $a_t$, transition behaviors.

To capture the multimodal sensor inputs required to sense a dense forest, we utilize the \acf{MVAE} model \citep{mvae}. This approach scales well to several high-dimensional input modalities  and gracefully handles prediction when modalities are missing, such as will occur each night for a robot using imagery from passive illumination. In our method, the \acf{MRSSM} combines these two recent advances, which are compatible through a modification of the reconstruction likelihood term in \acp{RSSM} training objective. We further extend the training-time sampling procedure of \acf{MVAE}, which provides its strong performance with missing modalities.%

\subsection{Recurrent state-space model}
\label{sec:planet}
Our work builds on the \ac{RSSM} introduced in PlaNet \citep{planet}, which combines a latent representation from a \ac{VAE} model with an \ac{RNN} to capture dynamics transitions. While several previous approaches combine a \ac{VAE} and an \ac{RNN} (e.g., World Models \citep{worldmodels} and \citep{8279425}), PlaNet has shown an increase in accuracy of prediction over longer prediction horizons by incorporating end-to-end training. 

The \ac{RSSM} consists of the following components:
\begin{enumerate}
    \item a transition model: $p(s_t|s_{t-1}, a_{t-1})$,
    \item an observation model: $p(o_t|s_t)$,
    \item the variational state distribution: $q(s_t | s_{t-1}, a_{t-1}, o_t)$.
\end{enumerate}
The transition model is implemented as an hybrid deterministic/stochastic \ac{RNN}, as described by \citet{planet}.
The observation model used to reconstruct images in the previous work was a transposed \ac{CNN}.
We will describe our extension to multimodal observations in Section \ref{sec_mrssm}. 
Training of each of the required models is performed in a unified procedure based on variational inference. Specifically, the desired parameters are those that maximize the likelihood of the observations gathered during training, under the time-series model. While this objective is intractable, a lower bound can be stated to guide optimization, as presented by \citet{planet}:
\begin{align}
\label{eq:elboplanet}
    &\ln p(o_{1:T}|a_{1:T-1}) = \ln \int \prod_{t=1}^{T} p(s_t | s_{t - 1}, a_{t - 1})p(o_{t}|s_t)ds_{1:T} \\
    &\geq \sum_{t=1}^{T} \Big( \e{q(s_t|o_{\leq t}, a_{< t})}{\ln p(o_t | s_t)}
    \nonumber \\
   &\hspace{10pt}- \longe{q(s_{t - 1}|o_{\leq t -1}, a_{< t -1})}{\mathcal{KL}\left[ q(s_t | o_{\leq t}, a_{< t}) || p(s_t | s_{t - 1}, a_{t - 1}) \right]}
    \Big) \\
    &:= \sum_{t=1}^{T} \text{ELBO}(o_t),
\end{align}
where $s_0$ is sampled from our initial state distribution, a Gaussian with mean 0 and variance 1.
From observed sequences of observations and actions, the parameters of each component model (neural network weights) can be optimized, leading to effective predictions of the future at test time.

\subsection{Multimodal variational autoencoder}
\label{sec:mvae}
Now, we need some way to map observations of different modalities to a common latent space.
This is non-trivial, especially when modalities can be missing or are not amenable to the same model architectures.
Instead of having single observation $o_t$ as in \autoref{sec:planet}, we now have $O_t = \{o_t^{1}, \dots, o_t^{N}\}$, a set containing $N$ sensor readings from the modalities present at time $t$.
The modalities present can change, for example due to sensor failure.
Methods which concatenate the output of two encoders \citep{travis} are inappropriate for this scenario.

\acp{MVAE} \citep{mvae} provide a way to encode multimodal data, while being robust to any combination of missing modalities.
Each modality has one encoder and decoder, which can have arbitrary architectures.
Each encoder of the modalities present at time $t$ separately outputs an estimate of the mean and variance for the latent variable $s_t$.
These estimates are then fused with what is called a product of experts \citep{poe}:
\begin{equation}\label{eq:poeprior}
    \hspace{-5pt} p(s_t|O_t, s_{t-1}, a_{t-1}) = \frac{1}{Z} p(s_t| s_{t-1}, a_{t-1})\prod_{o^i_t \in O_t} q(s_t|o^i_t)
\end{equation}
where $p(s_t| s_{t-1}, a_{t-1})$ is a prior coming from our \ac{RSSM} transition model, called the \textit{prior expert}, and $q(s_t|o^i_t)$ is the distribution outputted by the encoder for the modality $i$.
This assumes the independence of the different modalities conditional on the state. %
$Z$ is the partition function, or normalization constant, which is intractable in a general product of expert.
However, when our variational distribution $q$ and our prior is Gaussian, which is our case, we can compute the product's mean and variance in differentiable closed form \citep{mvae}.

Because the observations of different modalities are assumed to be conditionally independent given the states, we have that:
$p(O_t | s_t) = \prod_{o_t^i \in O_t} p(o_t^i|s_t)$ and
$\ln p(O_t | s_t) = \sum_{o_t^i \in O_t} \ln p(o_t^i|s_t)$.
In practice, $\sum_{o_t^i \in O_t} \lambda_i \ln p(o_t^i|s_t)$ is used in the \ac{ELBO} of the \ac{MVAE} to weight the reconstruction error of modalities differently \citep{mvae}; this amounts to having the factorization $p(O_t | s_t) = \prod_{o_t^i \in O_t} p(o_t^i|s_t)^{\lambda_i} / W$ where $W$ is a partition function which can be ignored when maximizing the \ac{ELBO}. Having low $\lambda_i$ leads to having a flat likelihood for modality $i$, assigning higher likelihood to reconstructions with high reconstruction error compared to high $\lambda_i$.

\subsection{Our approach: multimodal recurrent state space models}
\label{sec_mrssm}

\label{sec:system}
\begin{figure*}[h]
  \centering
  \vspace{5pt}
  \includegraphics[width=0.8\linewidth]{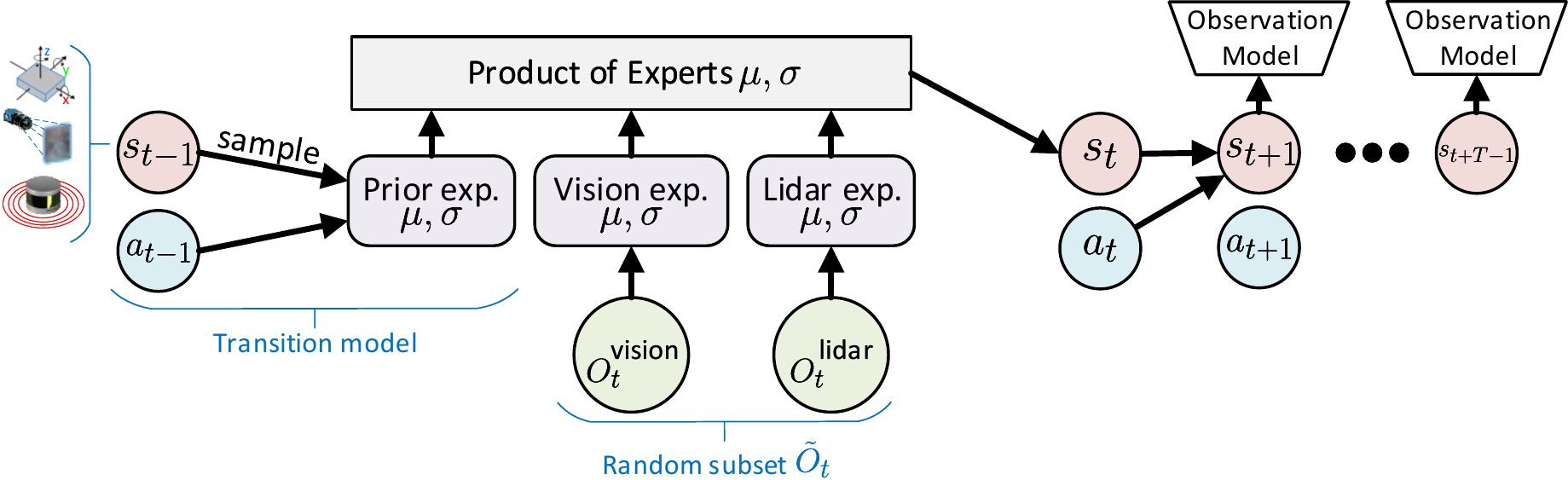}
  \caption{Diagram of how our model can do state estimation and perform predictions. In the filtering step, the model takes the previous posterior and action from time $t-1$ to generate a prior for time $t$. This prior is then fused with experts for the modalities present at time $t$ to generate a posterior. In the prediction step, one can take the posterior and action at time $t$ and run it through the transition model to get the prior for time $t+1$. To predict the future observations for time $t+1$, one can then sample from this prior and feed it to the observation model of the modalities one wants to predict. This prior can also be given to the transition model, along with a planned action, to perform predictions farther in the future,}
\label{fig:model}
\end{figure*}

As our goal is to predict outcomes for several steps from multimodal observations, we seek to combine \ac{RSSM} and \ac{MVAE} into a coherent approach. Noting that PlaNet already utilizes a \ac{VAE} to represent the observation model of a single mode, $p(o_t | s_t)$, the key for combining the methods is to properly integrate the multimodal observation likelihood into the \ac{RSSM} variational lower bound training objective.
The derivation is identical to \citep{planet}, just by substituting the observation likelihood $p(o_t|s_t)$ for the \ac{MVAE}'s observation likelihood of $p(O_t|s_t)$ \autoref{eq:poeprior}.

As suggested by \citet{mvae}, one needs to randomly remove modalities at training time to be able to train the individual experts and be robust to missing modalities.
An extension of what \citet{mvae} suggest to time series is randomly selecting a subset of present modalities, $\tilde{I}$, and creating a new time series, $\tilde{O}_t = \{o^i_t | i \in \tilde{I}, o^i_t \in O_t\}$ and maximizing ELBO($\tilde{O}_t$).
However, this does not take advantage of the fact that we have access to all information present at time $t$, $O_t$, because the modalities were randomly sub-sampled as part of the training procedure rather than being truly missing. It is interesting to train the model to be able to predict modalities that are not present in $\tilde{O}_t$.
For example, in our dynamics modeling application, it could be beneficial to train the vision expert to be able to predict velocities and accelerations from vision data alone, something that is not explicitly learnt by maximizing $\text{ELBO}(\tilde{O}_t)$. We can thus reformulate the \ac{ELBO} for $O$ instead of $\tilde{O}$:
\begin{align}
    \label{eq:newelbo}
    &\text{ELBO}(O_t) = \e{q(s_t|\tilde{O}_{\leq t}, a_{< t})}{
    \gamma_t
    \sum_{o^i_t \in O_t}\lambda_i\ln p(o^i_t | s_t)}
    \nonumber \\
    &\hspace{10pt}- \beta\longe{q(s_{t - 1}|\tilde{O}_{\leq t - 1}, a_{< t -1})}{\gamma_{t-1} \mathcal{KL}\left[ q(s_t | O_{\leq t}, a_{< t}) || p(s_t | s_{t - 1}, a_{t - 1}) \right]} \\
    &\hspace{43pt}:= \text{ELBO}(O_t|\tilde{O}_t)
\end{align}
where
\begin{equation}
    \gamma_t = \frac{q(s_t|O_{\leq t}, a_{< t})}{q(s_t|\tilde{O}_{\leq t}, a_{< t})}
\end{equation}
This shows that by doing importance sampling, we can express the \ac{ELBO} for $O_t$ using the product of experts for $\tilde{O}_t$.
We can get the posterior $q(s_t | O_{\leq t}, a_{< t})$ by always choosing all experts as our first subset, and then using it when computing the $\mathcal{KL}$-divergence term of the other sub-sampled $\tilde{O}_t$.
It makes sense, in an intuitive way, to use this posterior instead of the one coming from $\tilde{O}_t$.
Indeed, this $\mathcal{KL}$-divergence term is what enables our model to predict future states; by using the posterior from $O_t$, we are training our prior to be as close to the posterior containing the most information, no matter the information that was used to generate this prior.
Note that this works for computing the \ac{ELBO} of any other subset of $O_t$, but we restrict our attention to $O_t$ only.
When framed this way, training different combinations of experts then simply amounts to choosing different importance sampling distributions.
In practice, due to the high dimensionality of our latent state-space, we do not compute the importance sampling ratio for both expectations in the \ac{ELBO} and keep them to one, giving us a biased estimator of the \ac{ELBO}.
The approach, which we call a \acf{MRSSM}, is illustrated in \autoref{fig:model}.

\section{Experiments}
To verify that our model uses vision in a meaningful way, we first conduct controlled experiment in simulation.
We then move to more extensive real-world evaluation where we compare our new proposed \ac{ELBO} of \autoref{eq:newelbo} to \citep{mvae} and to two other baselines.
The experiments consists of building a model capable of predicting future velocities up to a chosen horizon $H$, given past observations $O_{\leq t}$ and intended actions $a_{<t+H}$.
Our code is based on an existing PyTorch implementation of PlaNet\footnote{https://github.com/Kaixhin/PlaNet} and \ac{MVAE}\footnote{https://github.com/mhw32/multimodal-vae-public} although it has been through substantial changes.

\subsection{Simulation experiments}
\label{sec:unreal}
\begin{figure}
    \centering
    \includegraphics[width=\linewidth]{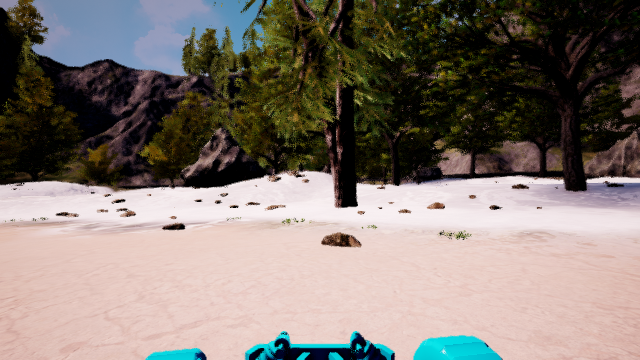}
    \caption{Sample image from our off-road robot simulator, where we can see the transition between terrain types, the gray rocky terrain having maximal traction and the white icy terrain having very low traction.
    Furthermore, small rocks visible here will affect the robot's dynamics, and bigger trees and rock add non-traversable obstacles.}
    \label{fig:unreal_sim}
    \vspace{-4ex}
\end{figure}
\begin{figure}
    \centering
    \vspace{5pt}
    \includegraphics[width=\linewidth]{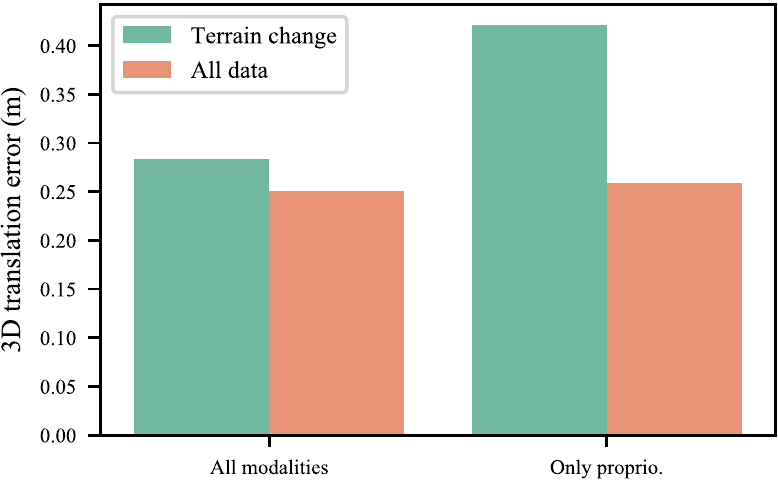}
    \caption{Median final pose error for 1s predictions, when the robot cross different terrain types and for the whole test dataset. For scale, the robot was going at an average speed of 5 m/s. We can see that when using all modalities including vision, performance degrades minimally when the traction situation changes. However, that is not the case when using only proprioception has the model has no access to information on the changing terrain. This shows that overall our model performs well when using only proprioception, but it is able to benefit from vision to predict traction changes.}
    \label{fig:results_unreal}
\end{figure}

We first validate our method using an off-road wheeled robot simulator based on Unreal Engine \citep{unrealengine}, shown in \autoref{fig:unreal_sim}, which gives rich visuals and realistic dynamics.
We test in simulation because data is hard to get and we can have full control of the terrains, as well as access to ground-truth.
We generated 4.38 hours of data in one map
using a random exploration policy, where the throttle and steering were sampled from a normal distribution centered at half throttle and centered steering respectively. 
The dataset is split into 684 trajectories, 10\% of which we keep as a validation dataset.
For simplicity, only three modalities were recorded: linear and angular velocity for proprioception, and a front-facing camera for exteroception.
The map has
three terrain types with different friction coefficients, enabling us to simulate situations where traction can change rapidly.
Furthermore, there are trees and rocks acting as rigid obstacles.
The simulation enables us to build a dataset where we know exactly when the robot's traction changes.
By using this information only for evaluation purposes, we can see the contribution of vision, where we expect it to help predict changes in traction, but not provide much benefit on flat terrain with constant traction.
The error metric we use is the Euclidean norm of the translation error of the predicted final pose, obtained by integrating the linear and angular velocity outputted by our model over the prediction horizon.
This metric allows us to evaluate both the angular and linear velocity predictions and how their errors accumulate with time.
The results of this basic experiment aiming to assess the influence of different modalities in varied scenario is visible in \autoref{fig:results_unreal}, where we isolate sequences where the robot transitions between different terrains. Our results show that using all modalities results in a 40\% decrease in translation error during terrain transitions, and even during non-transitions, our model using all modalities performs better than only proprioception sensing. 
We demonstrated that our robot is capable of predicting traction changes from visual information in an unsupervised manner with a controlled simulation experiment, now we quantify the performance of our method using a real-world dataset.

\subsection{Forest robot dynamics modeling}
\begin{figure}[ht!]
    \centering
    \vspace{5pt}
    \begin{subfigure}[b]{0.23\textwidth}
        \centering
        \includegraphics[width=\textwidth]{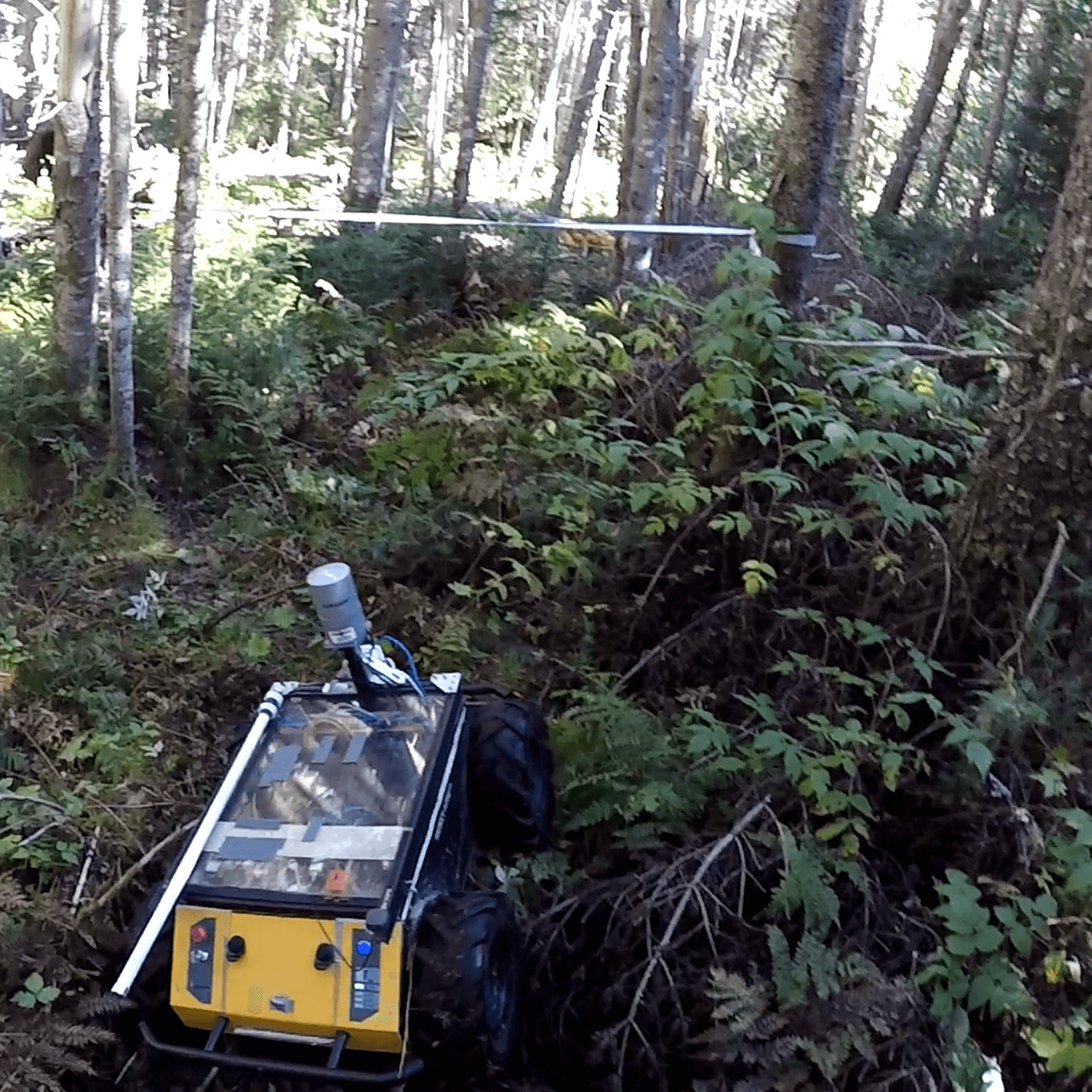}
        \caption{Branches}
    \end{subfigure}
    ~
    \begin{subfigure}[b]{0.23\textwidth}
        \centering
        \includegraphics[width=\textwidth]{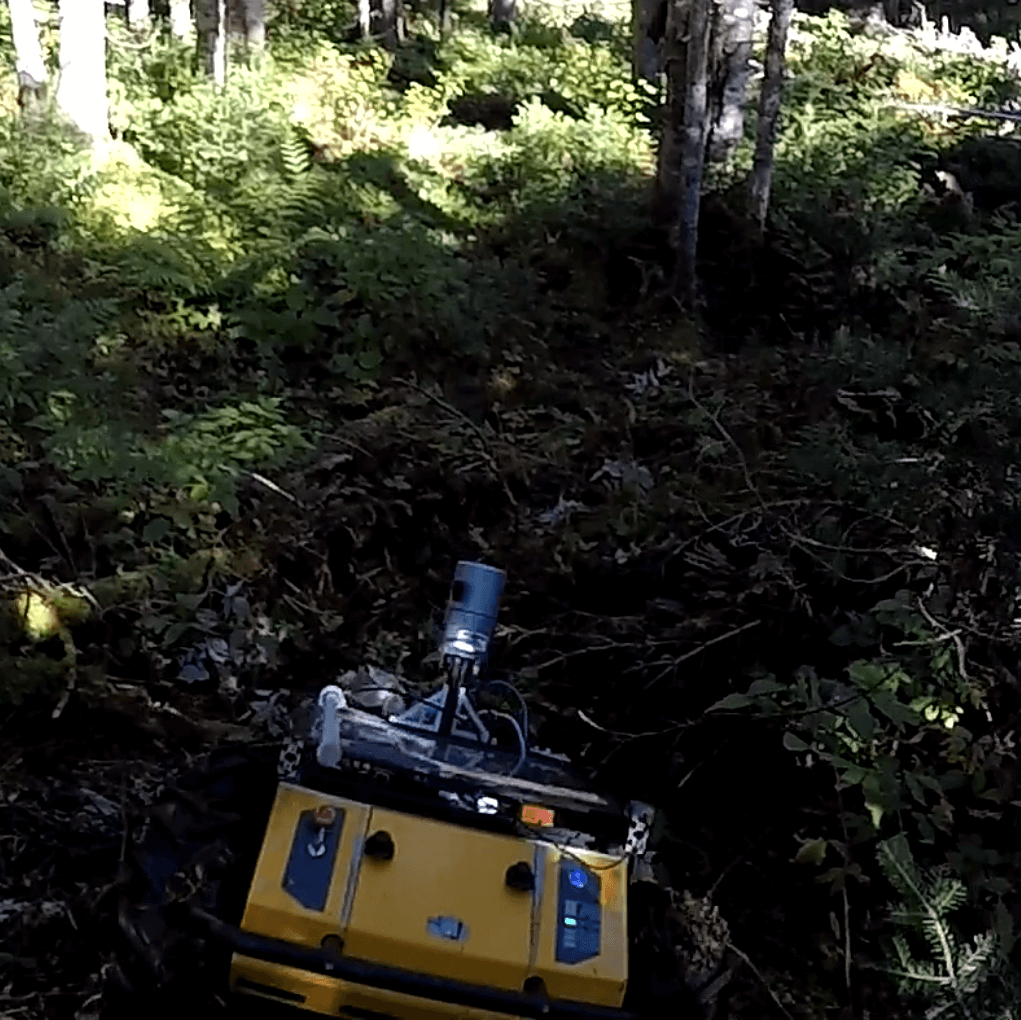}
        \caption{Downhill}
    \end{subfigure}
    \caption[]{Different events that affect the dynamics of the robot in our dataset. Branches can cause slippage or block the robot.
    Going up or downhill will also affect the dynamics through gravity.
    These images illustrate the diversity of scenario our method is being tested on, and the real-world conditions of the dataset used.}
    \label{fig:obstacles}
\end{figure}

Here we are working with the Montmorency dataset \citep{jfr}, consisting of 17 trajectories with high quality \ac{SLAM} based on off-line \ac{ICP}, allowing us to know the velocity of the robot.
These velocities are affected by the environment and contain four sites with different terrain types, causing varied effects on the dynamics.
This includes going over branches, loose moss and steep hill ascents/descents illustrated in \autoref{fig:obstacles}.
This dataset consists of visual, lidar and inertial data for a skid-steered robot with a maximal speed of 1m/s, along with the target forward and turning velocity given by a human driven operating the robot with a remote controller.
Thus, the five observation modalities with each their own expert are: linear velocity, angular velocity, tri-axial accelerometer, lidar height map of a 4 m\textsuperscript{2} area around the robot and the front camera view.
Both the lidar height map and front camera are given to our model with a $64 \times 64$ resolution, with the three channels for the lidar height map being the average $z$ coordinate of points falling in that pixel, the standard deviation of the height of these points and the average intensity of those points.
We used a fully connected encoder and observation model for the low-dimensional proprioceptive modalities, and a convolutional encoder and observation model for the vision and lidar.
The angular velocity was represented as a quaternion for the relative rotation in the last 0.1s (using the reconstruction loss proposed by \citet{Kendall_2017_CVPR}), which represents the discrete time-step used in our model at which all modalities are sampled.
One baseline we look at is assuming that the robot is always at its target velocity, called the \textit{Control} baseline.
For completeness, another baseline uses an \ac{RSSM} where the output of multiple encoders are concatenated; while this approach allows the fusion of multiple modalities, it is not compatible with missing modalities at test time.

\begin{figure}
    \centering
    \vspace{5pt}
    \includegraphics[width=\linewidth]{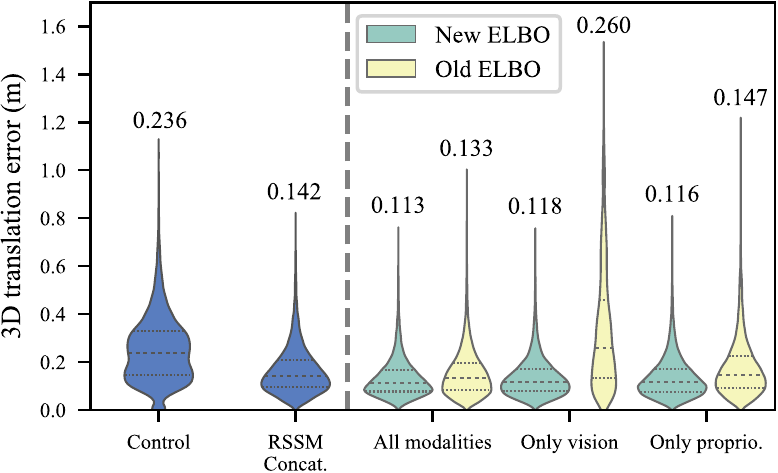}
    \caption{Error distribution for our baselines (left of dotted line) and our proposed \ac{MRSSM} under various modality ablations (right of dotted line) for the Montmorency dataset, with a prediction horizon of 3s.
    In each distribution, the first quartile, the median and the third quartile are drawn as dotted lines, and the median is displayed on top.
    One can see that the distribution of error, both in terms of median but also in terms of the tail of large errors, is lower for the \ac{MRSSM} with the new proposed \ac{ELBO}.
    Furthermore, the new proposed \ac{ELBO}
    performs better when modalities are removed.
    }
    \label{fig:results_montmorency}
\end{figure}

We use the same pose prediction error metric as in \autoref{sec:unreal}.
In \autoref{fig:results_montmorency}, we can see that our model beats the \textit{Control} baseline, even when doing predictions up to three seconds into the future.
To demonstrate the robustness to missing modalities of our approach, we remove different combination of modalities at test time, for the entire duration of the trajectory.
We test when all modalities are present, when only vision is present, and when only proprioception is present (accelerometer and velocity measurements).
Furthermore, we test both the \ac{ELBO} from \citep{mvae}, and the new proposed \ac{ELBO} from \autoref{eq:newelbo}.
We focus our evaluation on the site in a Maple grove, as this is the only site where we have more than 30 minutes of driving, with a total of 126 minutes in eight trajectories at this site.
Validating on trajectories on sites where there was only a few minutes of data present in the training set did not lead to conclusive results.
We perform validation in a leave-one-out manner, reporting \ac{RMSE} statistics on the eight validation trajectories.

Interestingly, when using the new \ac{ELBO} of \autoref{eq:newelbo} we could remove all modalities, including velocity measurements, and only keep vision.
Even in this scenario, our model is able to keep track of the robot's velocity (doing a form of deep visual odometry) and perform predictions that were better than the two baselines.
This was not the case for the previous \ac{ELBO} obtained from \citep{mvae}, where modality ablation lead to a much larger increase, doubling the median error.
Overall, for three seconds ahead pose predictions, our model leveraging all modalities with the \ac{ELBO} of \autoref{eq:newelbo} has the best average median error of 11.3cm, compared to 23.6cm for the \textit{Control} baseline.

\begin{figure}[h]
    \centering
    \vspace{5pt}
    \includegraphics[width=0.75\linewidth]{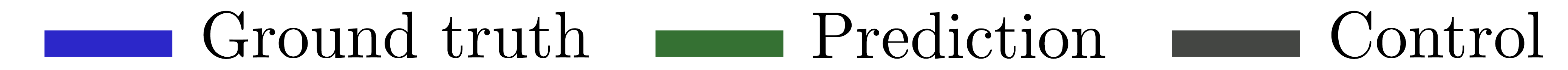}
    \includegraphics[width=\linewidth]{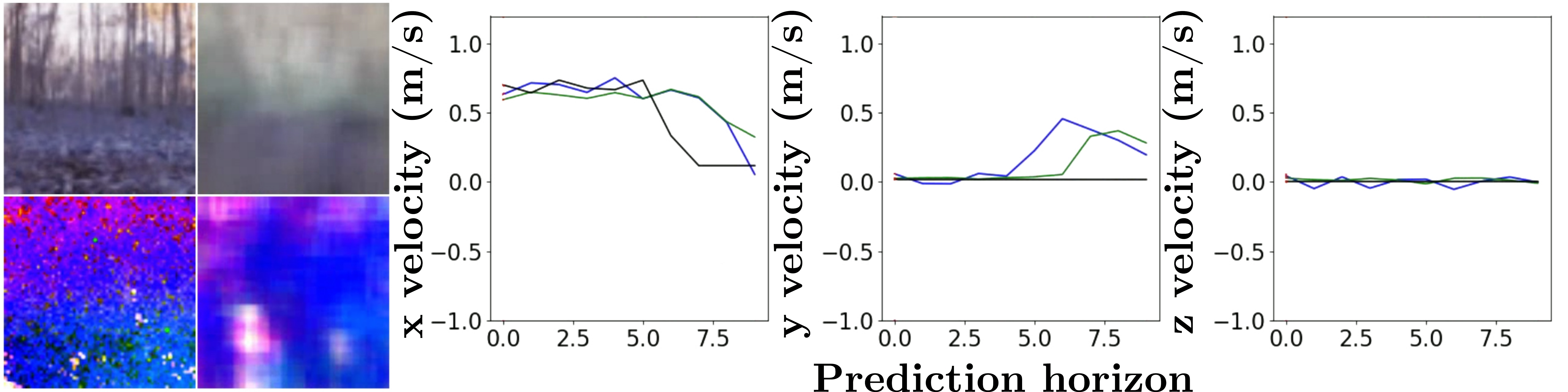}
    (a) Turning maneuver
    \includegraphics[width=\linewidth]{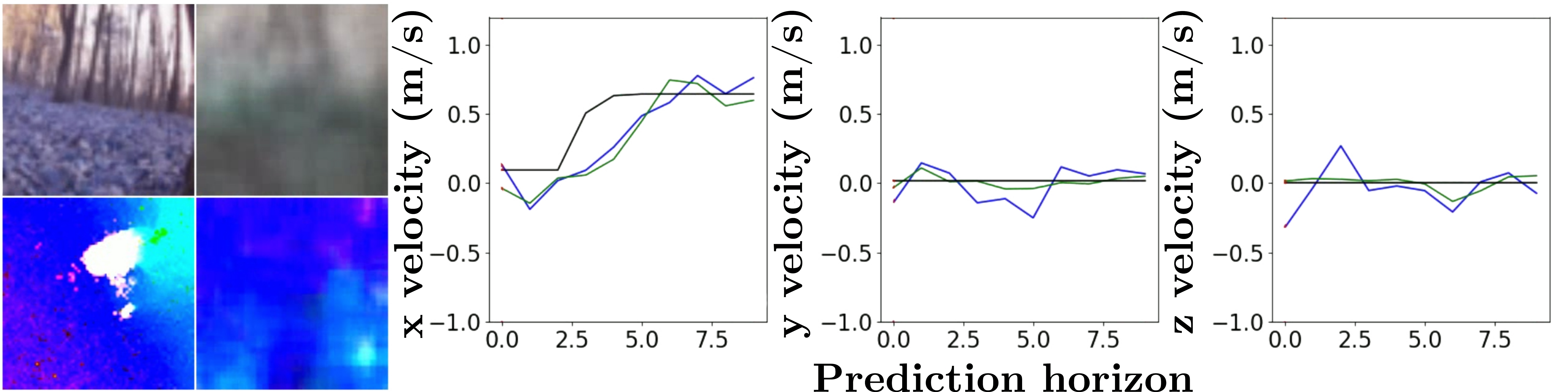}
    (b) Uneven terrain
    \includegraphics[width=\linewidth]{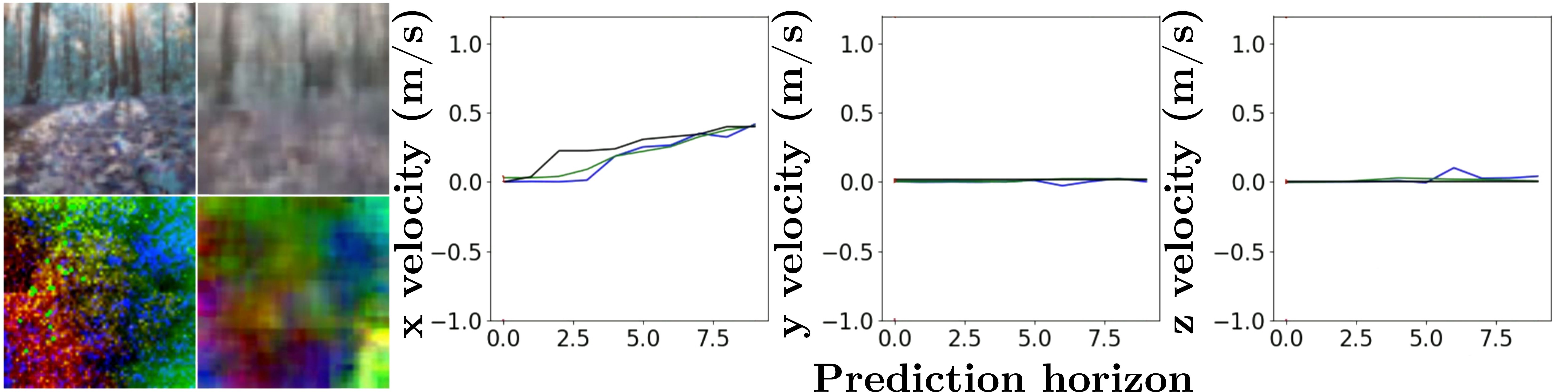}
    (c) Acceleration from stopped state
    \caption{Sample of predicted linear velocities and images, over an horizon of 1s, or 10 timesteps of 0.1s.
    Note how the visual prediction on the right, shown one timestep ahead of the current time, can become blurry for sequences that are more difficult to predict.
    One can also see the visual diversity even in the same testing site: (a) and (b) are in late fall with no leaves left on trees while (c) is in mid-October with colored leaves and a low sun vastly changing the visual appearance of the scene. This highlights the difficulty of learning from visual data in outdoor scenarios.}
    \label{fig:recons}
\end{figure}

Perhaps unsurprisingly, the \ac{RSSM} which concatenates the output of multiple encoders performed similarly to our approach using the \ac{ELBO} from \citep{mvae} when the latter has access to all modalities. However, both of these approaches were bested by our method using the new proposed \ac{ELBO}, showing that there is a gain in predictive power when learning to reconstruct missing modalities.
There was no modality ablation study for the concatenation baseline as it is unable to perform predictions when modalities are missing.
Altogether, our results demonstrate that the combination of the proposed \ac{MRSSM} and the \ac{ELBO} of \autoref{eq:newelbo} allows for 1) lower \ac{RMSE} of final pose predictions 2) robustness to missing modalities at test time.
Examples of predicted linear velocity and images for our approach with the new \ac{ELBO} are visible in \autoref{fig:recons}.

\section{Conclusion}
\label{sec:conclusion}
Driving off-road, across the wide spectrum of natural terrain requires integrating all available information.
In this paper, we have demonstrated that modern approaches for time-series modeling and learning multimodal representations are suitable for the problem of motion prediction in forests.
We derive a novel variational bound for learning time-series models using multimodal data that requires no human labeling, but still captures important environmental effects such as changing traction caused by different types.
Our quantitative results show that the motions of a four wheeled robot can be predicted accurately as it traverses a wide variety of slopes and contacts branches.
Our future work will include closing the control loop around our model, both by performing active exploration to collect informative data and by implementing planning and control using the valuable predictive power of our technique.

\printbibliography

\end{document}